\documentclass[10pt,twocolumn,letterpaper]{article}

\usepackage{cvpr}              

\usepackage{graphicx}
\usepackage{amsmath}
\usepackage{amssymb}
\usepackage{booktabs}

\usepackage[pagebackref,breaklinks,colorlinks]{hyperref}

\usepackage[capitalize]{cleveref}
\crefname{section}{Sec.}{Secs.}
\Crefname{section}{Section}{Sections}
\Crefname{table}{Table}{Tables}
\crefname{table}{Tab.}{Tabs.}

\def\cvprPaperID{*****} 
\def\confName{CVPR}
\def\confYear{2022}

\begin{document}

\title{UniUD Submission to the EPIC-Kitchens-100 Multi-Instance Retrieval Challenge 2023}

\author{Alex Falcon\\ University of Udine\\
{\tt\small falcon.alex@spes.uniud.it}
\and
Giuseppe Serra\\
University of Udine\\
{\tt\small giuseppe.serra@uniud.it}
}
\maketitle

\begin{abstract}
   In this report, we present the technical details of our submission to the EPIC-Kitchens-100 Multi-Instance Retrieval Challenge 2023. To participate in the challenge, we ensembled two models trained with two different loss functions on 25\% of the training data. Our submission, visible on the public leaderboard, obtains an average score of 56.81\% nDCG and 42.63\% mAP.
\end{abstract}

\section{Introduction\label{sec:intro}}
Text-video retrieval is a difficult task requiring a joint visual and textual understanding. The challenging Multi-Instance Retrieval benchmark offered by the EPIC-Kitchens-100 comprises around 70k egocentric video clips, each paired with a short instructional caption, which capture kitchen activities from different parts of the world. A major difference with respect to other video-language datasets is the possibility to assess the performance of the retrieval system with rank-aware metrics, such as nDCG and mAP. This makes it possible to truly assess the ranking capabilities of a model, which may highly differ from other metrics which are often used in retrieval contexts, e.g., recall metrics, as pointed out in \cite{wray2021semantic}. 

Our submission to the challenge consists of an ensemble of two solutions we trained on a random subset of 25\% of the training data. We opted to utilize a smaller dataset for training our solution due to the following reasons: firstly, using less training data enables lower energy consumption, aligning with the principles of 'green' AI; secondly, a data-efficient baseline is not only more accessible - as it can be trained using a single GPU in under an hour - but it also mirrors the human learning process, which involves extrapolating from relatively few examples. The first method included in our final solution uses a data augmentation technique on top of RANP \cite{falcon2022feature,falcon2022learning}, whereas the second one is a novel approach based on direct nDCG optimization \cite{pobrotyn2021neuralndcg}. The experiments show that, although using only 25\% of the training data, competitive results can be obtained. In particular, compared to the public leaderboard from last year, we obtain better nDCG (+1.5\%) than Satar et al.'s submission with less than 0.2\% difference in mAP, although they used the full training set \cite{satar2022exploiting}.

In Section \ref{sec:strategies} we provide details about the two optimization strategies \cite{falcon2022feature,pobrotyn2021neuralndcg} which we used. Section \ref{sec:exp} provides the implementation details regarding the network architecture, the ensemble strategy, and the results. Finally, we conclude the report in Section \ref{sec:conclusion}.

\section{Optimization strategies\label{sec:strategies}}
In this section, we provide an overview of the two strategies we used to perform the training.

\subsection{Feature-space multimodal data augmentation technique}
Data augmentation is typically performed to improve generalization by creating new samples through semantics-preserving transformations, e.g., random crops and horizontal flips in images. However, these transformations are commonly defined on the raw data, which may be difficult to obtain due to privacy or copyright issues (e.g., video datasets which are collected on YouTube). Moreover, applying these transformations on the raw data may lead to greater computational burdens, while also undermining the general applicability of the augmentation technique (e.g., a technique defined for images may require considerable efforts and reformulations to be applied to audio). To address these shortcomings, in \cite{falcon2022feature} we proposed a feature-space multimodal data augmentation technique which creates new samples by combining the latent representations of two semantically similar videos (or captions). We used the relevance function defined in \cite{Damen2021RESCALING} to determine the compatibility of two videos (or two captions) across those found in the training data under analysis. Then, we perform the augmentation for every training sample while also using RANP \cite{falcon2022learning} to improve the triplet loss both by selecting only irrelevant negatives and by additionally identifying relevant-positives different from the groundtruth.

\subsection{Direct optimization of nDCG}
The act of sorting a list of scores can be seen as a left-multiplication by a permutation matrix which is induced by the sorting operator itself. Therefore, by approximating the induced permutation matrix it is also possible to approximate the operator. NeuralSort proposes a continuous relaxation of the sorting operator which returns unimodal row-stochastic matrices in place of the permutation \cite{grover2019neuralsort}. Basing their work on NeuralSort, Pobrotyn et al. introduced NeuralNDCG, which is a loss function designed to directly optimize nDCG for learning-to-rank applications \cite{pobrotyn2021neuralndcg}. Specifically, the approximate matrix obtained via NeuralSort is left-multiplied to the groundtruth scores, which are obtained by applying the gain function to the groundtruth labels, then multiplied by the discount function and divided by the optimal DCG. In our implementation, we used the relevance values computed by the relevance function of \cite{Damen2021RESCALING} as the groundtruth labels.

\section{Experiments\label{sec:exp}}
This section concerns the base architecture and ensembling strategy, along the description of the experimental results.

\subsection{Implementation details}
\noindent\textbf{Base architecture and hyperparameters.} In both the considered models, HGR \cite{chen2020fine} was used as the base architecture. For each sentence, it builds a hierarchical structure of the semantic roles and then aggregates the textual features through graph message passing. Then, both global and local features are aligned with the corresponding visual features via a bidirectional loss. The training lasted 50 epochs, using the Adam optimizer with learning rate 1e-4, batch size 64, and a fixed margin set to 0.2. For the data-augmented model, we applied it on every sample (100\% augmentation chance) and used 0.15 as the threshold to distinguish relevant from irrelevant elements \cite{falcon2022learning}. To avoid ''leaks'' from the full training set, the augmentation is only performed among the samples found in the subset of size 25\%.

\noindent\textbf{Dataset.} We randomly sampled 25\% of the training set and used it as the training data. A non-overlapping small validation set was used to track the performance (the same used by, e.g., \cite{falcon2022learning}). The RGB, flow, and audio features extracted with TBN \cite{kazakos2019epic} and provided by the dataset authors were used for training.

\subsection{Ensemble strategy}
We deployed a simple ensemble strategy based on last year's submission \cite{falcon2022uniud}. First of all, we perform the training. Then, for each model, we extracted the similarity matrix computed on the test set. The two matrices are then summed and averaged: in this way, the similarity value for video $v_i$ and caption $q_j$ represents the mean of the two models. The final submission is done by using the mean similarity matrix. 

\subsection{Results}
Table \ref{tab:results} reports the performance obtained by the two single models and by the ensemble on the validation set. The ensemble (\textbf{Ens.}) obtained the highest performance and therefore was submitted to the leaderboard. When compared to the 2021 and 2022 editions of the EPIC-Kitchens-100 Multi-Instance Retrieval Challenge 2021 \cite{Damen2021CHALLENGES,Damen2022CHALLENGES}, we observe a better nDCG (+1.5\%) with similar mAP (only -0.2\%) than Satar et al.'s submission which obtained 55.33\% nDCG and 42.81\% mAP on average  \cite{satar2022exploiting}, and +3.3\% nDCG compared to the official JPoSE baseline and Hao et al.'s submission at the price of 1.4\% mAP \cite{wray2019fine,Damen2021CHALLENGES}. With respect to last year winners, Lin et al. and our previous submission \cite{falcon2022uniud,lin2022egocentric}, we obtain a good nDCG (around 4\% less) but the mAP is still far from optimal (around 7\% less than \cite{falcon2022uniud}). Noteworthily, all these results from previous works were obtained using the full training set, and not only 25\% of it.

\begin{table}[!t]
    \centering
    \begin{tabular}{c|ccc|ccc} \hline \hline
         & \multicolumn{6}{c}{Validation} \\ \hline
         & \multicolumn{3}{c|}{nDCG (\%)} & \multicolumn{3}{c}{mAP (\%)} \\ \hline 
        Mod. & v2t & t2v & avg & v2t & t2v & avg \\ \hline 
        Aug. & 70.6 & 70.7 & 70.6 & 64.5 & \textbf{62.9} & 63.7 \\ \hline 
        DOpt. & 70.2 & 71.0 & 70.6 & 64.0 & 58.9 & 61.5 \\ \hline 
        \textbf{Ens.} & \textbf{71.5} & \textbf{72.0} & \textbf{71.8} & \textbf{65.7} & 62.6 & \textbf{64.1} \\ \hline 
        \\ \hline 
         & \multicolumn{6}{c}{Official test} \\ \hline
         & \multicolumn{3}{c|}{nDCG (\%)} & \multicolumn{3}{c}{mAP (\%)} \\ \hline 
        \textbf{Ens.} & \textbf{58.65} & \textbf{54.96} & \textbf{56.81} & \textbf{47.79} & \textbf{37.48} & \textbf{42.63}  \\ \hline
        \hline 
    \end{tabular}
    \caption{Performance of the two single models (Aug.: feature-space multimodal data augmentation, DOpt.: direct optimization of nDCG) and the ensemble (Ens.) on the validation set (top) and test set (bottom) of EPIC-Kitchens-100.}
    \label{tab:results}
\end{table}

\section{Conclusion\label{sec:conclusion}} 
In this report, we summarized and briefly described our submission to the EPIC-Kitchens-100 Multi-Instance Retrieval Challenge 2023. The proposed method is an ensemble of the same base architecture trained with two different optimization strategies \cite{falcon2022feature,pobrotyn2021neuralndcg} on a random subset of 25\% of the original training data. The empirical results show that our solution achieves very good results compared to previous methods which used at least the full training data, e.g., we obtained better nDCG (+1.5\%) with similar mAP (-0.2\%) than last year's submission which ranked third. 

\section*{Acknowledgements}
This work was partly supported by the Italian Ministry of University and Research (MUR), within the project DM737\_HEU\_voucher\_2b\_FALCON (CUP G25F21003390007).

{\small
\bibliographystyle{ieee_fullname}
\bibliography{report}
}

\end{document}